\documentclass[11pt,a4paper]{article}
\pdfoutput=1
\usepackage[hyperref]{acl2019}
\usepackage{times}
\usepackage{latexsym}
\usepackage{url}

\aclfinalcopy

\usepackage{amsmath}
\usepackage{amssymb}
\usepackage{amsfonts}       
\usepackage{booktabs}
\usepackage{nicefrac}
\usepackage{microtype}
\usepackage{bm}
\usepackage{bbm}
\usepackage{mathtools}
\usepackage{enumitem}
\usepackage{subcaption}
\usepackage{tikz}

\usepackage{algorithm}
\usepackage{algorithmic}
\usepackage{natbib}
\usepackage{xspace}

\newcommand{\fair}{\textsuperscript{\normalfont 1}}
\newcommand{\ethz}{\textsuperscript{\normalfont 2}}

\author{Matt Le\fair \and
  Stephen Roller\fair \and
  Laetitia Papaxanthos\ethz \\
  {\bf Douwe Kiela}\fair \and 
  {\bf Maximilian Nickel}\fair \\
  \fair Facebook AI Research, New York, NY \\
  \ethz D-BSSE, ETH Z\"{u}rich, Switzerland
}
\date{}

\newcommand{\bless}{\textsc{Bless}}
\newcommand{\wbless}{\textsc{WBless}}
\newcommand{\bibless}{\textsc{BiBless}}
\newcommand{\hyperlex}{\textsc{Hyperlex}}
\newcommand{\shwartz}{\textsc{Shwartz}}
\newcommand{\leds}{\textsc{Leds}}
\newcommand{\eval}{\textsc{Eval}}
\renewcommand{\vec}[1]{\bm{#1}}

\newcommand{\vu}{\vec{u}}
\newcommand{\vv}{\vec{v}}
\newcommand{\vx}{\vec{x}}
\newcommand{\vy}{\vec{y}}

\newcommand{\lossfn}{\mathcal{L}}

\newcommand{\manifold}[1]{\mathbb{#1}}
\newcommand{\tansp}{\mathcal{T}}

\newcommand{\hyp}{\manifold{L}}
\newcommand{\R}{\mathbb{R}}

\newcommand{\ldot}[1]{\langle #1 \rangle_{\hyp}}
\newcommand{\lnorm}[1]{\| #1 \|_{\hyp}}
\newcommand{\proj}{\text{proj}}

\newcommand{\isa}{\texttt{is-a}\xspace}
\newcommand{\energy}{\mathcal{E}}
\newcommand{\hapert}{\alpha}
\newcommand{\hangle}{\phi}

\DeclareMathOperator\grad{grad}
\DeclareMathOperator*{\argmin}{arg\,min}
\newcommand{\modelpar}[1]{\paragraph{\normalfont\emph{#1}}}

\title{Inferring Concept Hierarchies \\ from Text Corpora via Hyperbolic Embeddings}
\begin{document}

\maketitle
\begin{abstract}
  We consider the task of inferring \isa relationships from large text corpora.
  For this purpose, we propose a new method combining hyperbolic embeddings and
  Hearst patterns. This approach allows us to set appropriate constraints for
  inferring concept hierarchies from distributional contexts while also being
  able to predict missing \isa-relationships and to correct wrong extractions.
  Moreover -- and in contrast with other methods -- the hierarchical nature of
  hyperbolic space allows us to learn highly efficient representations and to
  improve the taxonomic consistency of the inferred hierarchies. Experimentally,
  we show that our approach achieves state-of-the-art performance on several
  commonly-used benchmarks.
\end{abstract}

\section{Introduction}
\label{sec:intro}
Concept hierarchies, i.e., systems of \texttt{is-a} relationships, are
ubiquitous in knowledge representation and reasoning. For instance,
understanding \texttt{is-a} relationships between concepts is of special
interest in many scientific fields as it enables high-level abstractions for
reasoning and provides structural insights into the concept space of a domain.
A prime example is biology, in which taxonomies have a long history ranging from
\citet{linnaeus1758systema} up to recent efforts such as the Gene Ontology and
the UniProt
taxonomy~\citep{ashburner2000gene,gene2016expansion,apweiler2004uniprot}.
Similarly, in medicine, ontologies like MeSH and ICD-10 are used to organize
medical concepts such as diseases, drugs, and
treatments~\citep{rogers1963medical,simms1992icd}.
In Artificial Intelligence, concept hierarchies provide valuable information for
a wide range of tasks such as automated reasoning, few-shot learning, transfer
learning, textual entailment, and semantic
similarity~\citep{resnik1993selection,lin:1998:icml,kgs/berners2001semantic,dagan2010recognizing,bowman:2015:emnlp,zamir2018taskonomy}.
In addition, \isa relationships are the basis of complex knowledge graphs such
as \textsc{DBpedia}~\citep{kgs/auer2007dbpedia} and
\textsc{Yago}~\citep{kgs/suchanek2007yago,kgs/hoffart2013yago2} which have found
important applications in text understanding and question
answering

\begin{figure}[t]
  \centering
  \includegraphics[width=\columnwidth]{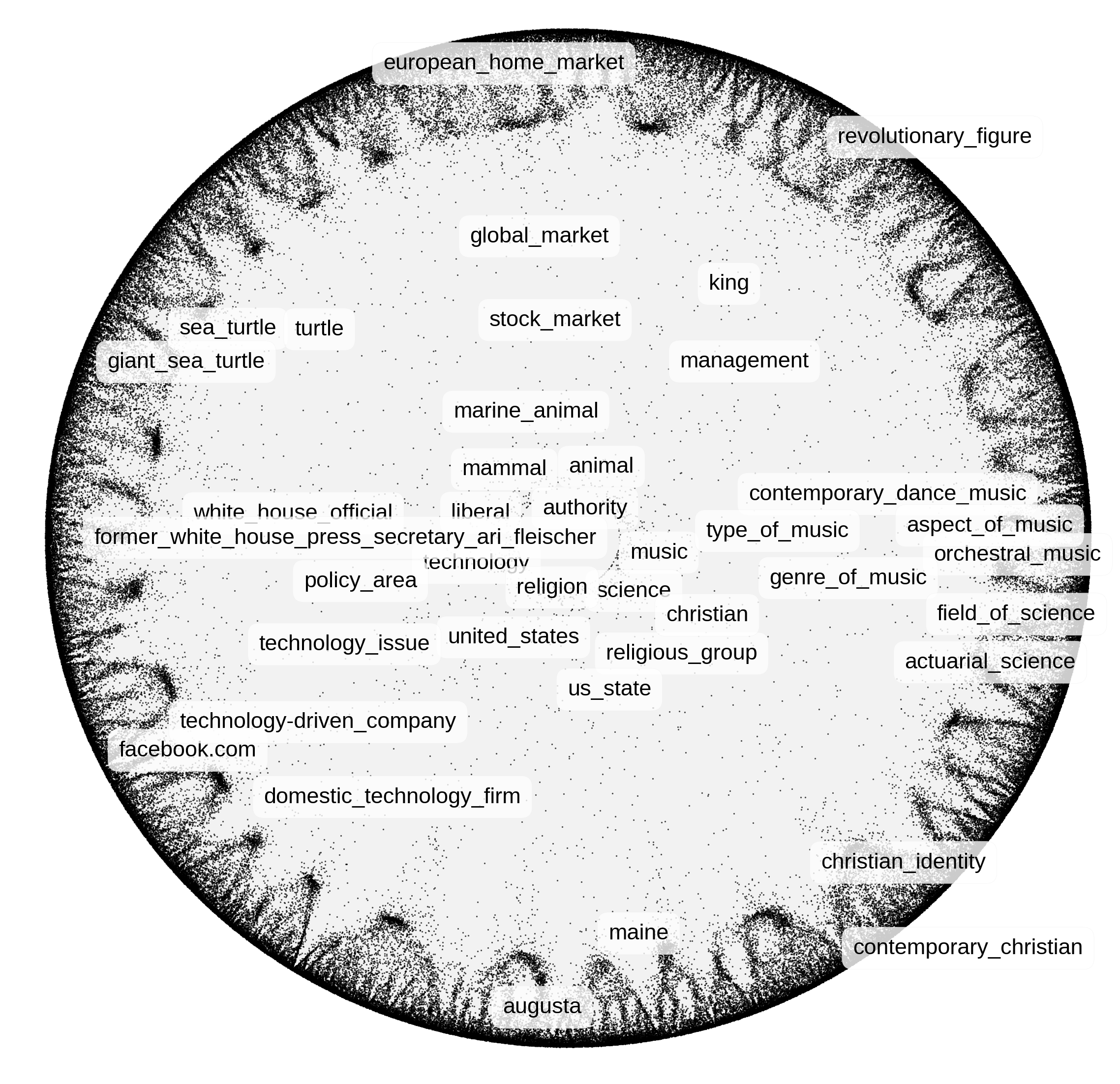}
  \caption{Example of a two-dimensional hyperbolic embedding of the extracted Hearst Graph.}\label{fig:embedding}
\end{figure}

Creating and inferring concept hierarchies has, for these reasons, been a long
standing task in fields such as natural language processing, the semantic web,
and artificial intelligence.
Early approaches such as
\textsc{WordNet}~\citep{miller1990introduction,miller1998wordnet} and
\textsc{CyC}~\citep{kgs/lenat1995cyc} have focused on the \emph{manual}
construction of high-quality ontologies.
To increase scalability and coverage, the focus in recent efforts such as
\textsc{ProBase}~\citep{kgs/wu2012probase} and
\textsc{WebIsaDB}~\citep{kgs/seitner2016webisadb} has shifted towards
\emph{automated} construction.

\begin{figure}[t]
\centering
\includegraphics[width=.8\linewidth]{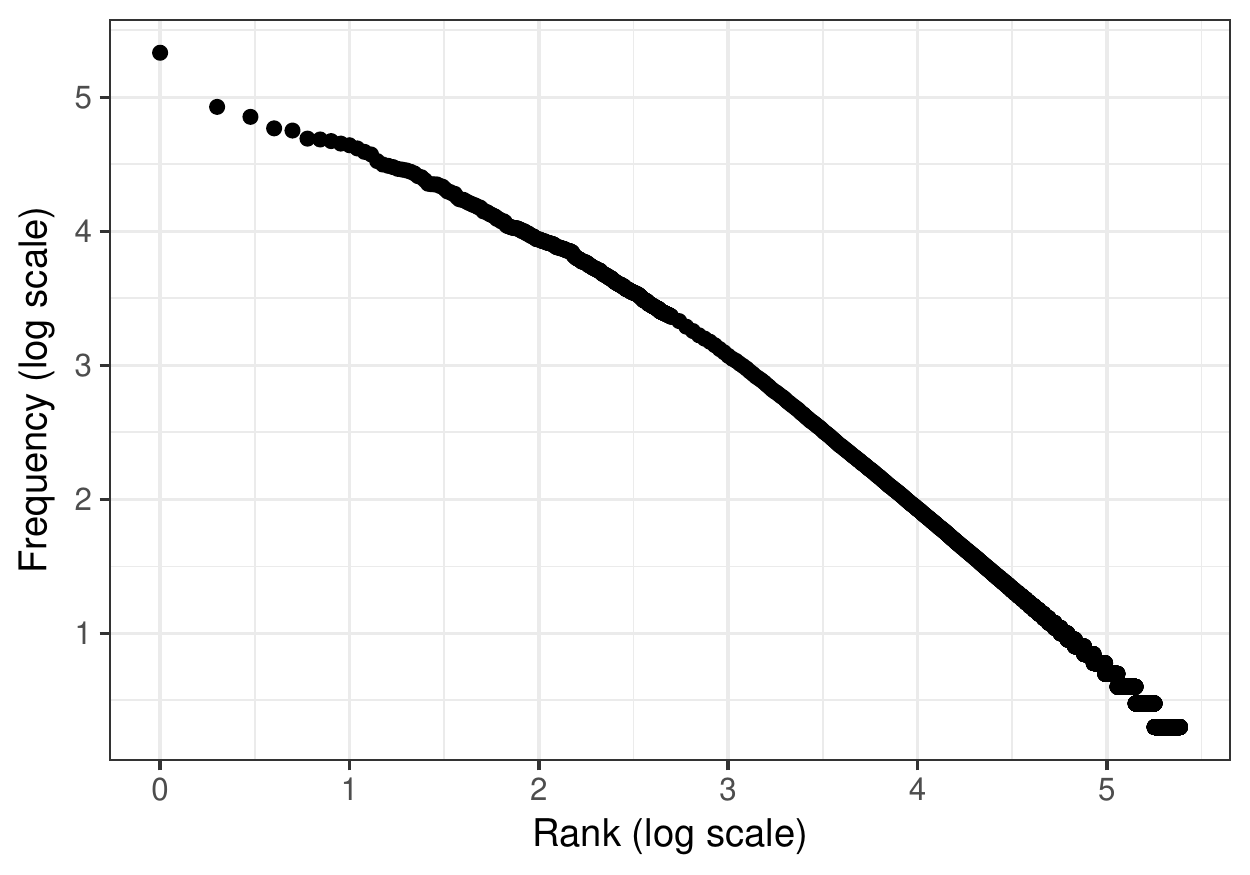}
\caption{\label{fig:dist}
Frequency distribution of words appearing in the Hearst pattern corpus (on a
log-log scale).}
\end{figure}

In this work, we consider the task of inferring concept hierarchies from large
text corpora in an \emph{unsupervised} way. For this purpose, we combine Hearst
patterns with recently introduced hyperbolic
embeddings~\citep{mine/nickel2017poincare,mine/nickel2018learning}, what
provides important advantages for this task. First, as
\citet{mine/roller2018hearst} showed recently, Hearst patterns provide important
constraints for hypernymy extraction from distributional contexts.
However, it is also well-known that Hearst patterns are prone to missing and
wrong extractions, as words must co-occur in exactly the right pattern to be
detected successfully.
For this reason, we first extract potential \isa relationships from a corpus
using Hearst patterns and build a directed weighted graph from these
extractions. We then embed this Heart Graph in hyperbolic space to infer missing
hypernymy relations and remove wrong extractions.
By using hyperbolic space for the embedding, we can exploit the following
important advantages:

\begin{description}[leftmargin=1.5em]
\item[Consistency] Hyperbolic entailment cones
  \citep{hyperbolic/ganea2018hyperbolic} allow us to enforce transitivity of
  \isa-relations in the entire embedding space. This improves the taxonomic
  consistency of the model, as it enforces that $(x, \isa, z)$ if $(x, \isa, y)$
  and $(y, \isa, z)$. To improve optimization properties, we also propose a new
  method to compute hyperbolic entailment cones in the Lorentz model of
  hyperbolic space.
  
\item[Efficiency] Hyperbolic space allows for very low dimensional embeddings of
  graphs with latent hierarchies and heavy-tailed degree distributions. To embed
  large Hearst graphs -- which exhibit both properties (e.g., see
  Figure~\ref{fig:dist}) -- this is an important advantage. In our experiments, we
  will show that hyperbolic embeddings allow us to decrease the embedding
  dimension by over an order of magnitude while outperforming SVD-based methods.
  
\item[Interpretability] In hyperbolic embeddings, similarity is captured via
  distance while hierarchy is captured through the norm of embeddings. In
  addition to semantic similarity this allows us to get additional insights from
  the embedding such as the generality of terms.
\end{description}

Figure~\ref{fig:embedding} shows an example of a two-dimensional embedding of the
Hearst graph that we use in our experiments. Although we will use higher
dimensionalities for our final embedding, the visualization serves as a good
illustration of the hierarchical structure that is obtained through the
embedding.



\section{Related Work}
\label{sec:related}
\paragraph{Hypernym detection}
Detecting \isa-relations from text is a long-standing task in natural language
processing. A popular approach is to exploit high-precision
\emph{lexico-syntactic patterns} as first proposed
by~\citet{hearst1992automatic}. These patterns may be predefined or learned
automatically \citep{snow2005learning,shwartz2016improving}. However, it is well
known that such pattern-based methods suffer significantly from missing
extractions as terms must occur in exactly the right configuration to be
detected \cite{shwartz2016improving,mine/roller2018hearst}. Recent works improve
coverage by leveraging search engines~\citep{kozareva2010semi} or by exploiting
web-scale corpora \citep{kgs/seitner2016webisadb}; but also come with
significant precision trade-offs.

To overcome the sparse extractions of pattern-based methods, focus has recently
shifted to distributional approaches which provide rich representations of
lexical meaning. These methods alleviate the sparsity issue but also require
specialized similarity measures to distinguish different lexical relationships.
To date, most measures are inspired by the \emph{Distributional Inclusion
  Hypothesis} (DIH; \citealt{geffet2005distributional}) which hypothesizes that
for a subsumption relation (\emph{cat}, \texttt{is-a}, \emph{mammal}) the
subordinate term (\emph{cat}) should appear in a subset of the contexts in which
the superior term (\emph{mammal}) occurs. Unsupervised methods for hypernymy
detection based on distributional approaches include \emph{WeedsPrec}
\citep{weeds2004characterising}, \emph{invCL} \citep{lenci2012identifying},
\emph{SLQS} \cite{santus2014chasing}, and \emph{DIVE}
\cite{chang2018distributional}. Distributional representations that are based on
positional or dependecy-based contexts may also capture crude
Hearst-pattern-like features \citep{levy2015supervised,roller2016relations}.
\citet{shwartz2017hypernyms} showed that such contexts plays an important role
for the success of distributional methods.

Recently, \citet{mine/roller2018hearst} performed a systematic study of
unsupervised distributional and pattern-based approaches. Their results showed
that pattern-based methods are able to outperform DIH-based methods on several
challenging hypernymy benchmarks. Key aspects to good performance where the
extraction of patterns from large text corpora and using embedding methods to
overcome the sparsity issue. Our work builds on these findings by replacing
their embeddings with ones with a naturally hierarchical geometry.

\paragraph{Taxonomy induction}
Although detecting hypernymy relationships is an important and difficult task,
these systems alone do not produce rich taxonomic graph structures
\cite{camacho2017why}, and complete taxonomy induction may be seen as a parallel
and complementary task.

Many works in this area consider a taxonomic graph as the starting point, and
consider a variety of methods for growing or discovering areas of the graph. For
example, \citet{snow2006semantic} train a classifier to predict the likelihood
of an edge in WordNet, and suggest new undiscovered edges, while
\citet{kozareva2010semi} propose an algorithm which repeatedly crawls for new
edges using a web search engine and an initial seed taxonomy.
\citet{cimiano2005learning} considered learning ontologies using Formal Concept
Analysis. Similar works consider noisy graphs discovered from Hearst patterns,
and provide algorithms for pruning edges until a strict hierarchy remains
\citep{velardi2005evaluation,kozareva2010semi,velardi2013ontolearn}.
\citet{maedche2001ontology} proposed a method to learn ontologies in a Semantic
Web context.

\paragraph{Embeddings} Recently, works have proposed a variety of graph
embedding techniques for representing and recovering hierarchical structure.
Order-embeddings \cite{embeddings/vendrov2016order} represent text and images
with embeddings, where the ordering over individual dimensions form a partially
ordered set. Hyperbolic embeddings treat words as points in non-Euclidean
geometries, and may be viewed as a continuous generalization of tree structures
\cite{mine/nickel2017poincare,mine/nickel2018learning}. Extensions have
considered how distributional co-occurrences may be used to augment order
embeddings \cite{li2017improved} and Hyperbolic embeddings
\cite{dhingra2018embedding}. Other recent works have focused on the often
complex overlapping structure of word classes, and induced hierarchies using
box-lattice structures \cite{vilnis2018probabilistic} and Gaussian word
embeddings \cite{athiwaratkun2018hierarchical}. Compared to many of the purely
graph-based works, these methods generally require extensive supervision of
hierarchical structure, and cannot learn taxonomies using only unstructured
noisy data. Recently, \citet{tifrea2018poincar} proposed an extension of
\textsc{GloVe}~\citep{pennington2014glove} to hyperbolic space. Our experimental
results in Section~\ref{sec:exps}, which show substantial gains over the results
reported by \citet{tifrea2018poincar} for hypernymy prediction, underline the
importance of selecting the right distributional context for this task.


\section{Methods}
\label{sec:org2da8a80}
In the following, we discuss our method for unsupervised learning of concept
hierarchies. We first discuss the extraction and construction of the Hearst
graph, followed by a description of the Hyperbolic Embeddings.

\subsection{Hearst Graph}
\label{sec:org2593680}

\begin{table}[t]
  \small
  \centering
  \begin{tabular}{l}
    \toprule
    \textbf{Pattern}\\
    \midrule
    X which is a (example \(\vert{}\) class \(\vert{}\) kind \(\vert{}\) \ldots) of Y\\
    X (and \(\vert{}\) or) (any \(\vert{}\) some) other Y\\
    X which is called Y\\
    X is \texttt{JJS} (most)? Y\\
    X a special case of Y\\
    X is an Y that\\
    X is a !(member \(\vert{}\) part \(\vert{}\) given) Y\\
    !(features \(\vert{}\) properties) Y such as X\(_{\text{1}}\), X\(_{\text{2}}\), \ldots\\
    (Unlike \(\vert{}\) like) (most \(\vert{}\) all \(\vert{}\) any \(\vert{}\) other) Y, X\\
    Y including X\(_{\text{1}}\), X\(_{\text{2}}\), \ldots\\
    \bottomrule
  \end{tabular}
  \caption{\label{tab:patterns} Hearst patterns used in this study. Patterns are
    lemmatized, but listed as inflected for clarity.}
  \centering
\end{table}

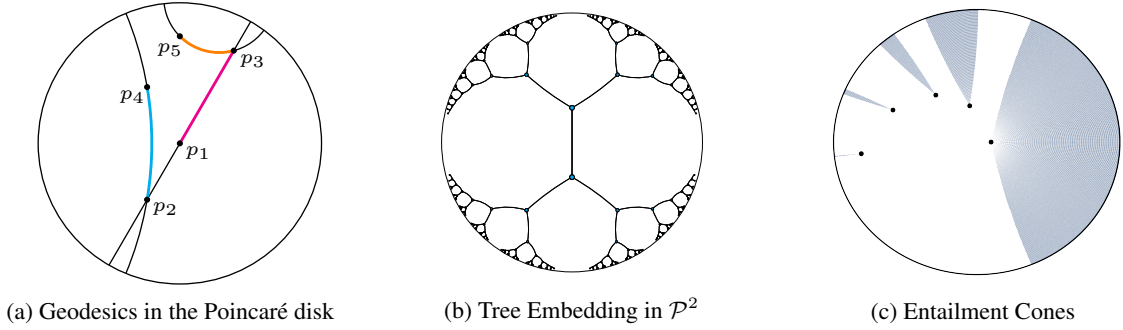
\begin{figure*}[t]
  \centering
  \begin{minipage}{0.33\linewidth}
    \centering
    \resizebox{0.8\linewidth}{!}{\pgfdeclarelayer{background}
\pgfdeclarelayer{foreground}
\pgfsetlayers{background,main,foreground}
\begin{tikzpicture}[very thick,scale=0.5]
\begin{pgfonlayer}{foreground}\draw (0,0) circle (3.0);\end{pgfonlayer}
\tikzstyle{segment}=[line width=0.3mm]

\begin{pgfonlayer}{background}
\end{pgfonlayer}

\draw[black] (-1.142, 2.774) arc (22.383:-22.383:7.285);
\draw[black] (-0.333, 2.981) arc (-173.619:-36.381:1.175);
\draw[black] (-1.500, -2.598) -- (1.500, 2.598);
\draw[segment,magenta] (0.000, 0.000) -- (1.142, 1.979);
\draw[segment,orange] (0.000, 2.285) arc (-135.230:-74.770:1.175);
\draw[segment,cyan] (-0.693, 1.201) arc (9.486:-9.486:7.285);

\begin{pgfonlayer}{foreground}
\draw[fill=black,black] (0.000,0.000) circle (0.05);
\draw[fill=black,black] (-0.693,-1.201) circle (0.05);
\draw[fill=black,black] (1.142,1.979) circle (0.05);
\draw[fill=black,black] (-0.693,1.201) circle (0.05);
\draw[fill=black,black] (0.000,2.285) circle (0.05);
\node at (0.4,-0.2) {\tiny $p_1$};
\node at (-0.3,-1.4) {\tiny $p_2$};
\node at (1.55,1.77) {\tiny $p_3$};
\node at (-1,1) {\tiny $p_4$};
\node at (-0.2,2) {\tiny $p_5$};
\end{pgfonlayer}
\end{tikzpicture}}
    \subcaption{Geodesics in the Poincar\'{e} disk}\label{fig:geodesics}
  \end{minipage}%
  \begin{minipage}{0.33\linewidth}
    \centering
    \includegraphics[width=.7\columnwidth]{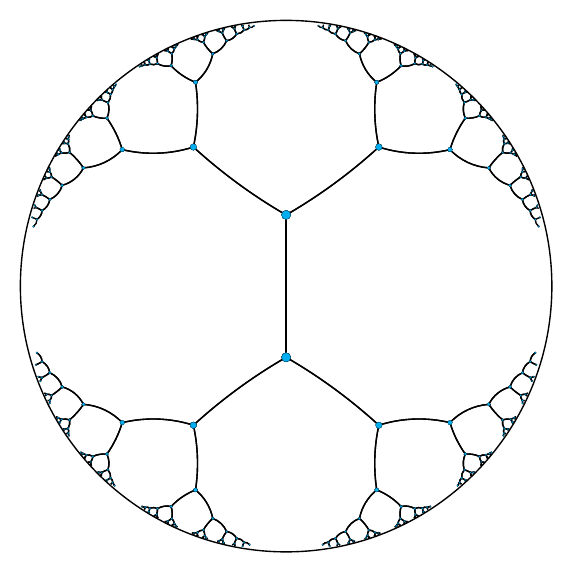}
    \subcaption{Tree Embedding in $\mathcal{P}^2$}\label{fig:tree_embedding}
  \end{minipage}%
  \begin{minipage}{0.33\linewidth}
    \centering
    \includegraphics[width=.77\columnwidth]{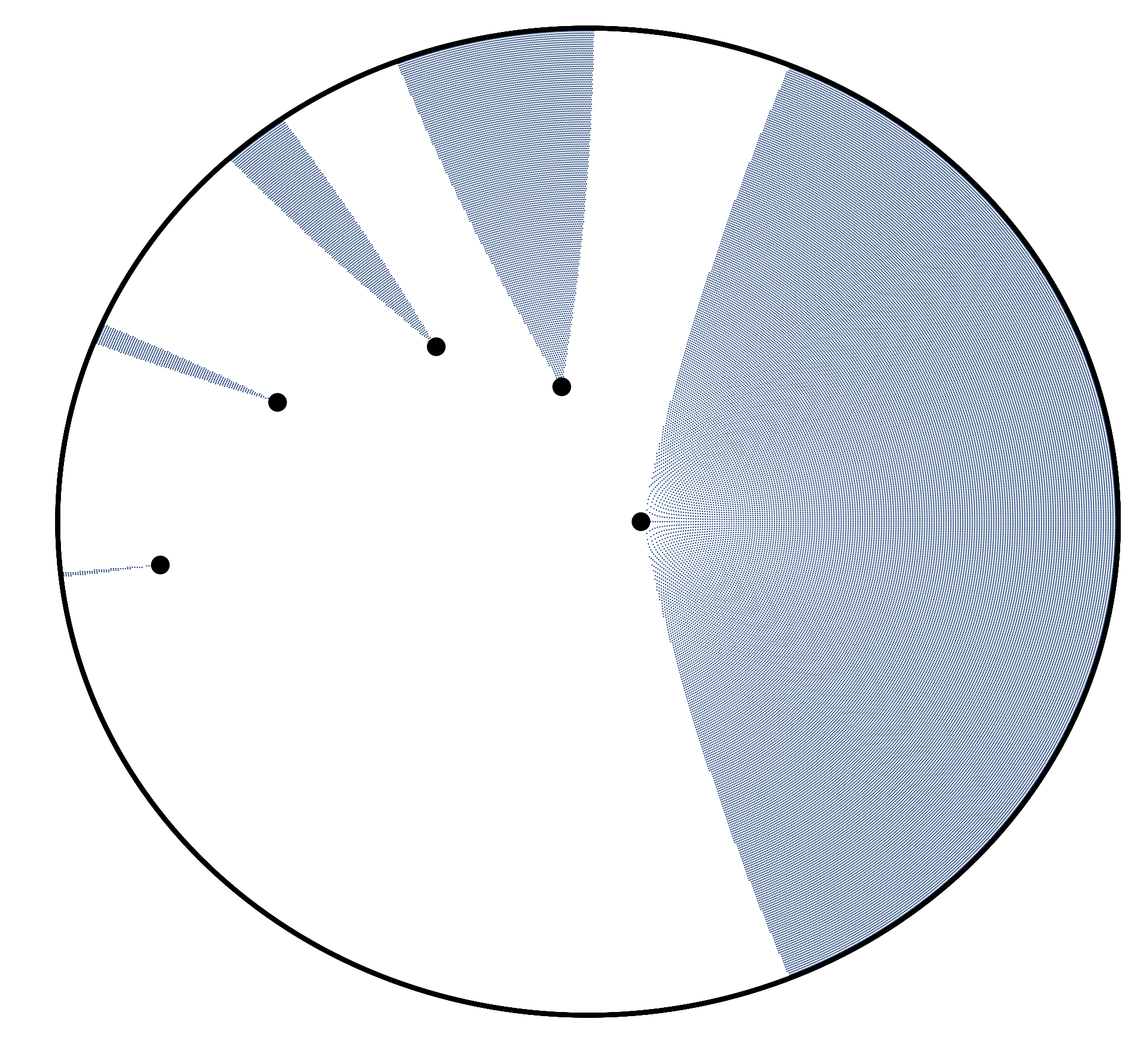}
    \subcaption{Entailment Cones}\label{fig:econes}
  \end{minipage}
  \caption{\subref{fig:geodesics}) Geodesics in the Poincar\'{e} disk model of
    hyperbolic space. Geodesics between points are arcs that are perpendicular
    to the boundary of the disk. For curved arcs, midpoints are closer to the
    origin of the disk ($p_1$) than the associated points, e.g. ($p_3$, $p_5$).
    \subref{fig:econes}) Entailment cones for different points in
    $\manifold{P}^2$.}
\end{figure*}

The main idea introduced by \citet{hearst1992automatic} is to exploit certain
\emph{lexico-syntactic patterns} to detect \texttt{is-a} relationships in
natural language. For instance, patterns like ``\texttt{NP}\(_y\) \emph{such as}
\texttt{NP}\(_x\)'' or ``\texttt{NP}\(_x\) \emph{and other} \texttt{NP}\(_y\)''
often indicate a hypernymy relationship $(u, \isa, v)$. By treating unique noun
phrases as nodes in a large, directed graph, we may construct a \emph{Hearst
  Graph} using only unstructured text and very limited prior knowledge in form
of patterns. Table~\ref{tab:patterns} lists the only patterns that we use in this
work.
Formally, let \({E = \{(u, v)\}_{i=1}^N}\) denote the set of \isa relationships
that have been extracted from a text corpus. Furthermore, let \(w(u,v)\) denote
how often we have extracted the relationship \(u, \isa, v\). We then represent
the extracted patterns as a weighted directed graph \(G = (V,E,w)\) where \(V\)
is the set of all extracted terms.

Hearst patterns afford a number of important advantages in terms of data
acquisition: they are embarrassingly parallel across both sentences and distinct
Hearst patterns, and counts are easily aggregated in any MapReduce setting
\cite{dean2004mapreduce}. Our own experiments, and those of
\citet{kgs/seitner2016webisadb}, demonstrate that this approach can be scaled to
large corpora such as
\textsc{CommonCrawl}.\footnote{\url{http://commoncrawl.org}} As
\citet{mine/roller2018hearst} showed, pattern matches also provide important
contextual constraints which boost signal compared to methods based on the
Distributional Inclusion Hypothesis.

However, na\"ively using Hearst pattens can easily result in a graph that is
extremely sparse: pattern matches naturally follow a long-tailed distribution
that is skewed by the occurrence probabilities of constituent words (see
Figure~\ref{fig:dist}) and many true relationships are unlikely to ever appear in a
corpus (e.g. ``long-tailed macaque \texttt{is-a} entity''). This may be
alleviated with generous, low precision patterns \cite{kgs/seitner2016webisadb},
but the resulting graph will contain many false positives, inconsistencies, and
cycles. For example, our own Hearst graph contains the cycle: (\emph{area},
\texttt{is-a}, \emph{spot}), (\emph{spot}, \emph{commercial}),
(\emph{commercial}, \emph{promotion}), (\emph{promotion}, \emph{area}), which is
caused by the polysemy of \emph{spot} (location, advertisement) and \emph{area}
(location, topical area).


\subsection{Hyperbolic Embeddings}
\label{sec:hyperbolic}
\citet{mine/roller2018hearst} showed that low-rank embedding methods, such as
Singular Value Decomposition (SVD), alleviate the aforementioned sparsity issues
but still produce cyclic and inconsistent predictions. In the following, we will
discuss how hyberbolic embeddings allow us to improve consistency via strong
hierarchical priors in the geometry.

First, we will briefly review necessary concepts of hyperbolic embeddings: In
contrast to Euclidean or Spherical space, there exist multiple equivalent models
for hyperbolic space.\footnote{e.g., the Poincar\'{e} ball, the Lorentz model, the
  Poincar\'{e} upper half plane, and the Beltrami-Klein model} Since there exist
transformations between these models that preserve all geometric properties
(including isometry), we can choose whichever is best suited for a given task.
In the following, we will first discuss hyperbolic embeddings based on the
Poincar\'{e}-ball model, which is defined as follows: The Poincar\'{e}-ball model is the
Riemannian manifold ${\manifold{P}^n = (\manifold{B}^n, d_p)}$, where
$\manifold{B}^n = \{\vx \in \R^n : \|\vx\| < 1\}$ is the \emph{open}
$n$-dimensional unit ball and where $d_p$ is the distance function
\begin{align}
d_p(\vu, \vv) & = \cosh^{-1} \left(1 + 2 \delta(\vu, \vv) \label{eq:pdist} \right) \\
\delta(\vu, \vv) & = \frac{\|\vu - \vv\|^2}{(1 - \|\vu\|^2)(1 - \|\vv\|^2)} \notag
\end{align}

Hyperbolic space has a natural hierarchical structure and, intuitively, can be
thought of as a continuous versions of trees.
This property becomes evident in the Poincar\'{e} ball model: it can be seen from
Equation~\ref{eq:pdist}, that the distance within the Poincar\'{e} ball changes smoothly
with respect to the norm of a point $\vu$. Points that are close to the origin
of the disc are relatively close to all other points in the ball, while points
that are close to the boundary are relatively far apart\footnote{This can be
  seen by considering how the Euclidean distance in $\delta(\vu,\vv)$ is scaled
  by the norms of the respective points}. This locality property of the distance
is key for learning continuous embeddings of hierarchical structures and
corresponds to the behavior of shortest-paths in trees.

\paragraph{Hyperbolic Entailment Cones}
\citet{hyperbolic/ganea2018hyperbolic} used the hierarchical properties of
hyperbolic space to define entailment cones in an embedding. The main idea of
hyperbolic entailment cones (HECs) is to define for each possible point $\vv$ in
the space, an entailment region in the form of a hyperbolic cone
$\mathcal{C}_{\vv}$. Points $\vu$ that are located inside a cone
$\mathcal{C}_{\vv}$ are assumed to be children of $\vv$. The width of each cone
is determined by the norm $\|\vv\|$ of the associated base point. The closer
$\vv$ is to the origin, i.e., the more general the basepoint is, the larger the
width of $\mathcal{C}_{\vv}$ becomes and the more points are subsumed in the
entailment cone. Equation~\ref{fig:econes} shows entailment cones for different points
in $\manifold{P}^2$.
To model the possible entailment $(u, \isa, v)$, we can then use the
\emph{energy} function
\begin{equation}
  \energy(u, v) = \max(0, \hangle(\vu, \vv) - \hapert(\vv))
  \label{eq:energy}
\end{equation}

In Equation~\ref{eq:energy}, $\hapert(\vv)$ denotes the half-aperture of the cone
associated with point $\vv$ and $\hangle(\vv, \vu)$ denotes the angle between
the half-lines $(\vv\vu$ and $(0\vv$.
If $u \in \mathcal{C}_{\vv}$, i.e., if the angle between $\vu$ and $\vv$ is
smaller than the half aperture of $\mathcal{C}_{\vv}$, it holds that $\energy(u,
v) = 0$. If $u \not \in \mathcal{C}_{\vv}$, the energy $\energy(u, v)$ can then
be interpreted as smallest angle of a rotation bringing $\vu$ into the cone
associated with $\vv$.

Given a Hearst graph $G = (V, E, w)$, we then compute embeddings of all terms in
the following way: Let \(\vv_i\) denote the embedding of term \(i\) and let
$\Theta = \{\vv_i\}_{i=1}^M$. To minimize the overall energy of the embedding,
we solve the optimization problem
\begin{equation}
  \hat{\Theta} = \argmin_{\Theta \in \manifold{H}^n} \sum_{u,v\, \in V}\lossfn(u,v)
  \label{eq:loss-optim}
\end{equation}
where
\begin{equation*}
  \lossfn(u,v) = \begin{cases}
    \energy(u, v) & \text{if } (u,v) \in E \\
    \max(0, \gamma - \energy(u, v)) & \text{otherwise}
  \end{cases}
\end{equation*}
The goal of Equation~\ref{eq:loss-optim} is therefore to find a joint embedding of all
terms that best explains all observed Hearst patterns.

\begin{figure}[tb]
  \centering
  \includegraphics[trim={2cm 7cm 2cm
    7cm},clip,width=.8\columnwidth]{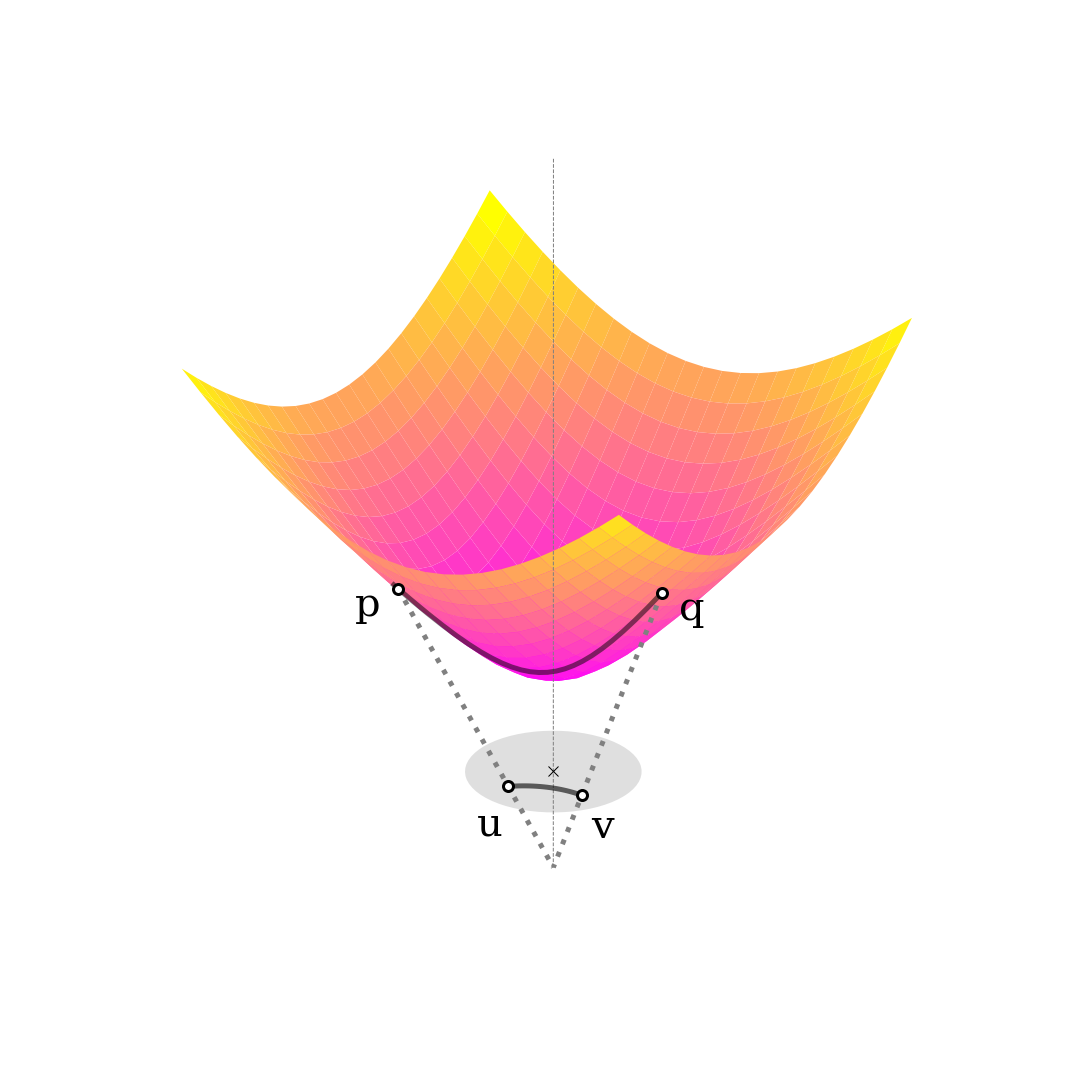}
  \caption{Mapping between $\manifold{P}$ and $\hyp$. Points ($p$, $q$)
    lie on the surface of the upper sheet of a two-sheeted hyperboloid.
    Points ($u$, $v$) are the mapping of ($p$, $q$) onto the Poincar\'{e} disk using
    Equation~\ref{eq:p-to-h}.}
  \label{fig:hyperboloid}
\end{figure}
 
\paragraph{Lorentz Entailment Cones}
The optimization problem Equation~\ref{eq:loss-optim} is agnostic of the hyperbolic
manifold on which the optimization is performed.
\citet{hyperbolic/ganea2018hyperbolic} developed hyperbolic entailment cones in
the Poincar\'{e}-ball model.
However, as \citet{mine/nickel2018learning} pointed out, the Poincar\'{e}-ball
model is not optimal from an optimization perspective as it is prone to
numerical instabilities when points approach the boundary of the ball. Instead,
\citet{mine/nickel2018learning} proposed to perform optimization in the Lorentz
model and use the Poincar\'{e} ball only for analysis and visualization.
Here, we follow this approach and develop entailment cones in the Lorentz model
of hyperbolic space. The Lorentz model is defined as follows: let \(\vx\), \(\vy
\in \R^{n+1}\) and let
\begin{equation*}
  \ldot{\vx, \vy} = -x_0 y_0 + \sum_{i=1}^n x_n y_n
\end{equation*}
denote the \emph{Lorentzian scalar product}. The Lorentz model of
\(n\)-dimensional hyperbolic space is then the Riemannian
manifold ${\hyp^n = (\manifold{H}^n, d_\ell)}$, where 
\begin{equation*}
  \manifold{H}^n = \{\vx \in \R^{n+1} : \ldot{\vx, \vx} = -1, x_0 > 0\}
\end{equation*}
denotes the upper sheet of a two-sheeted \(n\)-dimensional hyperboloid and where
the associated distance function on $\hyp$ is given as
\begin{equation*}
  d_\ell(\vx, \vy) = \cosh^{-1}(-\ldot{\vx, \vy}) \label{eq:ldist}
\end{equation*}

Due to the equivalence of both models, we can define a mapping between both
spaces that preserves all properties including isometry. Points in the Lorentz
model can be mapped into the Poincar\'{e} ball via the diffeomorphism \(p :
\manifold{H}^n \to \manifold{P}^n\), where
\begin{equation}
  p(x_0, x_1, \ldots, x_n) = \frac{(x_1, \ldots, x_n)}{x_0 + 1}
  \label{eq:p-to-h}
\end{equation}
Furthermore, points in \(\manifold{P}^n\) can be mapped to \(\manifold{L}^n\) via
\begin{equation*}
  p^{-1}(x_1, \ldots, x_n) = \frac{(1 + \|x\|^2, 2x_1,\ldots,2x_n)}{1 - \|x\|^2}
\end{equation*}
See also Figure~\ref{fig:hyperboloid} for an illustration of the Lorentz model
and its connections to the Poincar\'{e} ball.

To define entailment cones in the Lorentz model, it is necessary to derive
$\hapert(\vv)$ and $\hangle(\vu,\vv)$ in $\hyp^n$. Both quantities can be derived
easily from the hyperbolic law of cosines and the mapping between
$\manifold{P}^n$ and $\hyp^n$. In particular, it holds that 

\begin{align*}
  \hapert(\vv) & = \sin^{-1} \left(2K / (-v_0 - 1) \right)\\
  \hangle(\vu,\vv) & = \cos^{-1}\left( \frac{v_0 + u_0 \ldot{\vu,\vv}}{\|\vv^\prime\|\sqrt{\ldot{\vu,\vv}^2 - 1}} \right)
\end{align*}

Due to space restrictions, we refer to the supplementary material for the full derivation.

\paragraph{Training} To solve Equation~\ref{eq:loss-optim}, we follow
\citet{mine/nickel2018learning} and perform stochastic optimization via
Riemannian SGD (RSGD; \citealt{optimization/bonnabel2013stochastic}). In RSGD,
updates to the parameters \(\theta\) are computed via
\begin{equation}
  \theta_{t+1} = \exp_{\theta_t}(-\eta \grad f(\theta_t))
  \label{eq:sgd}
\end{equation}
where \(\grad f (\theta_t) \) denotes the
\emph{Riemannian gradient} and \(\eta\) denotes the learning rate.
In Equation~\ref{eq:sgd}, the Riemannian gradient of \(f\) at \(\theta\) is computed via
\begin{equation*}
  \grad f(\theta_t) = \proj_{\theta_t} \left(g_\ell^{-1}\nabla f(\theta) \right)
  \label{eq:direction}
\end{equation*}
where $\nabla f(\theta)$ denotes the \textit{Euclidean} gradient of $f$ and
where
\begin{align*}
  \proj_{\vx}(\vu) & = \vu + \ldot{\vx, \vu} \vx \\
  g_\ell^{-1}(\vx) & = \operatorname{diag}([-1, 1, \ldots, 1])
\end{align*}
denote the projection from the ambient space $\R^{n+1}$ onto the tangent space
$\tansp_{\vx}\hyp^n$ and the inverse of the metric tensor, respectively.
Finally, the exponential map for $\hyp^n$ is computed via
\begin{equation*}
  \exp_{\vx}(\vv) = \cosh(\lnorm{\vv})\vx + \sinh(\lnorm{\vv})\frac{\vv}{\lnorm{\vv}}
\end{equation*}
where \(\lnorm{\vv} = \sqrt{\ldot{\vv,\vv}}\) and \(\vv \in \tansp_{\vx} \hyp^n\).

As suggested by \citet{mine/nickel2018learning}, we initialize the embeddings
close to the origin of \(\manifold{L}^n\) by sampling from the uniform
distribution \(\mathcal{U}(-0.001, 0.001)\) and by setting \(x_0\) to $\sqrt{1 + ||x'||^2}$, 
what ensures that the sampled points are located on the surface of the hyperboloid.



\section{Experiments}
\label{sec:exps}
To evaluate the efficacy of our method, we evaluate on several commonly-used
hypernymy benchmarks (as described in \cite{mine/roller2018hearst}) as well as
in a reconstruction setting (as described in \cite{mine/nickel2017poincare}).
Following~\citet{mine/roller2018hearst}, we compare to the following methods for
unsupervised hypernymy detection:

\begin{table*}[t]
  \small
  \begin{center}
    \begin{tabular}{lccccccccc}
      \toprule
      & \multicolumn{5}{c}{{\bf Detection} (AP)}  & \multicolumn{3}{c}{{\bf Direction} (Acc.)}  & {\bf Graded} ($\rho$) \\
      \cmidrule(r){2-6} \cmidrule(lr){7-9} \cmidrule(l){10-10}
      &  {\bless} & \eval &  \leds &  {\shwartz} &  {\wbless} & \bless & \wbless & \bibless & \hyperlex\\
      \midrule
      Cosine      &    .12 &    .29 &    .71 &     .31 &    .53 &    .00 &     .54 &      .52 &    .14 \\
      WeedsPrec   &    .19 &    .39 &    .87 &     .43 &    .68 &    .63 &     .59 &      .45 &    .43 \\
      invCL       &    .18 &    .37 &{\bf.89}&     .38 &    .66 &    .64 &     .60 &      .47 &    .43 \\
      SLQS        &    .15 &    .35 &    .60 &     .38 &    .69 &    .75 &     .67 &      .51 &    .16 \\
      \midrule
      p(x, y)     &    .49 &    .38 &    .71 &     .29 &    .74 &    .46 &     .69 &      .62 &{\bf.62}\\
      ppmi(x, y)  &    .45 &    .36 &    .70 &     .28 &    .72 &    .46 &     .68 &      .61 &    .60\\
      sp(x, y)    &    .66 &    .45 &    .81 &     .41 &    .91 & {\bf .96} &     .84 &      .80 &    .51\\
      spmi(x, y)  &.76 & .48 &    .84 & .44 & .96 & {\bf .96} &.87 & .85 &    .53\\ 
      \midrule
      HypeCones & {\bf .81} &   {\bf .50} &   {\bf .89} &      {\bf .50} &     {\bf .98} & .94 &     {\bf .90} &      {\bf .87} & .59 \\
      \bottomrule
    \end{tabular}
  \caption{Experimental results comparing distributional and pattern-based
    methods in all settings.}
  \label{tab:results}
  \end{center}
\end{table*}

\paragraph{Pattern-Based Models}
Let $E = \{(x, y)\}_{i=1}^N$ be the set of Hearst patterns in our corpus, $w(x,
y)$ be the count of how many times $(x, y)$ occurs in
$E$, and $W = \sum_{(x, y) \in E} w(x, y)$. We then consider the following
pattern-based methods:

\modelpar{Count Model (p)} This models simply outputs the count, or
equivalently, the extraction probabilities of Hearst patterns, i.e.,
\[
  \text{p}(x, y) = \frac{w(x, y)}{W}
\]

\modelpar{PPMI Model (ppmi)}
To correct for skewed occurrence probabilities, the PPMI model predicts
hypernymy relations based on the Pointwise Mutual Information over the Hearst
pattern corpus. Let $p^{-}(x) = \Sigma_{(x, y) \in \mathcal{P}} w(x, y) / W$ and
${p^+(x) = \Sigma_{(y, x) \in E} w(y, x)}$, then:
\[
  \text{ppmi}(x, y) = max\left(0, log \frac{p(x, y)}{p^-(x)p^+(y)} \right)
\]

\modelpar{SVD Count (sp)}
To account for missing relations, we also compare against low-rank embeddings of the
Hearst corpus using Singular Value Decomposition (SVD). Specifically, let $X \in
R^{M x M}$, such that $X_{ij} = w(i, j) / W$ and $U \Sigma V^{\top}$ be the
singular value decomposition of $X$, then:
\[
  \text{sp}(x, y) = u_x^{\top} \Sigma_r v_y
\]

\modelpar{SVD PPMI (spmi)} We also evaluate against the SVD of the PPMI matrix,
which is identical to $\text{sp}(i,j)$, with the exception that $X_{ij} =
\text{ppmi}(i, j)$, instead of $p(i, j)$. \citet{mine/roller2018hearst} showed
that this method provides state-of-the-art results for unsupervised
hypernymy detection.

\modelpar{Hyperbolic Embeddings (HypeCone)} We embed the Hearst graph into hyperbolic space
as described in section \ref{sec:hyperbolic}. At evaluation time, we predict the
likelihood using the model energy $\energy(u, v)$.

\paragraph{Distributional Models}
The distributional models in our evaluation are based on the DIH, i.e., the
idea that contexts in which a narrow term $x$ may appear (ex: \emph{cat}) should be a
subset of the contexts in which a broader term $y$ (ex: \emph{animal}) may appear.

\modelpar{WeedsPrec} The first distributional model we consider is
\textit{WeedsPrec} \cite{weeds2004characterising}, which captures the features
of $x$ which are \textit{included} in the set of more general term's features,
$y$:
\[
  \text{WeedsPrec}(x, y) = \frac{\sum_{i=1}^n x_i \cdot \mathbbm{1}_{y_i > 0}}{\sum_{i=1}^n x_i}
\]

\modelpar{invCL} \citet{lenci2012identifying}, introduce the idea of
distributional \text{exclusion} by also measuring the degree to which the broader
term contains contexts not used by the narrower term. The degree of
\textit{inclusion} is denoted as:
\[
  \text{CL}(x, y) = \frac{\sum_{i=1}^n \min(x_i, y_i)}{\sum_{i=1}^n x_i}
\]
To measure the inclusion of $x$ and $y$ and the non-inclusion of $y$ in $x$,
invCL is then computed as
\[
  \text{invCL}(x, y) = \sqrt{\text{CL}(x, y) \cdot (1 - \text{CL}(y, x)}
\]

\modelpar{SLQS} The SLQS model is based on the informativeness hypothesis
\cite{santus2014chasing,shwartz2017hypernyms}, i.e., the idea that general words
appear mostly in uninformative contexts, as measured by entropy. SLQS depends on
the median entropy of a term's top $k$ contexts:
\[
  E_x = \text{median}_{i=1}^k [H(c_i)]
\]
where $H(c_i)$ is the Shannon entropy of context $c_i$ across all terms. SLQS is
then defined as:
\[
  \text{SLQS}(x, y) = 1 - E_x / E_y
\]

\paragraph{Corpora and Preprocessing}
We construct our Hearst graph using the same data, patterns, and procedure as
described in \cite{mine/roller2018hearst}: Hearst patterns are extracted from
the concatenation of GigaWord and Wikipedia. The corpus is tokenized,
lemmatized, and POS-tagged using CoreNLP 3.8.0 \cite{manning2014stanford}. The
full set of Hearst patterns is provided in Table~\ref{tab:patterns}. These include
prototypical Hearst patterns, like ``animals [such~as] big cats,'' as well as
broader patterns like ``New Year [is the most important] holiday.'' Noun phrases
were allowed to match limited modifiers, and produced additional hits for the
head of the noun phrase. The final corpus contains circa 4.5M matched pairs,
431K unique pairs, and 243K unique terms.

\begin{table*}
  \small
  \centering
  \begin{tabular}{llllllllll}
    \toprule
    & \multicolumn{3}{c}{\bf Animals} &  \multicolumn{3}{c}{\bf Plants} &  \multicolumn{3}{c}{\bf Vehicles} \\
    \cmidrule(r){2-4}\cmidrule(lr){5-7}\cmidrule(l){8-10}
    & All & Missing & Transitive & All & Missing & Transitive & All & Missing & Transitive \\
    \midrule
    p(x, y)                        &   350.18 & 512.28 & 455.27 &  271.38 & 393.98 & 363.73 &     43.12 & 82.57 & 66.10 \\
    ppmi(x, y)                       &   350.47 & 512.28 & 455.38 &  271.40 & 393.98 & 363.76 &    43.20 & 82.57 & 66.16 \\
    sp(x, y)                     &    56.56 & 77.10 & 11.22 &   43.40 & 64.70 & 17.88 &     9.19 & 26.98 & 14.84\\
    spmi(x, y)                    &    58.40 & 102.56 & 12.37 &   40.61 & 71.81 & 14.80 &      9.62 & 17.96 & 3.03 \\
    \midrule
    HypeCones & {\bf 25.33} & {\bf 37.60} & {\bf 4.37} & {\bf 17.00} & {\bf 31.53} & {\bf 6.36} &      {\bf 5.12} & {\bf 10.28} & {\bf 2.74} \\
    $\Delta \%$ & 56.6 & 51.2 & 61.1 & 58.1 & 51.3 & 57.0 & 44.3 & 42.8 & 9.6 \\
    \bottomrule
  \end{tabular}
  \caption{Reconstruction of Animals, Plants, and Vehicles subtrees in
    \textsc{WordNet}.}\label{tab:hovy}
\end{table*}

\paragraph{Hypernymy Tasks} We consider three distinct subtasks for evaluating
the performance of these models for hypernymy prediction:
\begin{itemize}
\item \emph{Detection}: Given a pair of words ($u$, $v$), determine if $v$ is a hypernym of $u$.  
\item \emph{Direction}: Given a pair ($u$, $v$), determine if $u$ is more general than $v$ or vise versa.
\item \emph{Graded Entailment}: Given a pair of words ($u$, $v$), determine the
  degree to which $u$ is a $v$.
\end{itemize}
For detection, we evaluate all models on five commonly-used benchmark datasets:
{\bless}~\cite{baroni2011gems}, {\leds}~\cite{baroni2012leds},
{\eval}~\cite{santus2015evalution}, {\shwartz}~\cite{shwartz2016improving}, and
{\wbless}~\cite{weeds2014learning}, In addition to positive hypernymy relations,
these datasets includes negative samples in the form of
random pairs, co-hyponymy, antonymy, meronymy, and adjectival relations. For
directionality and graded entailment, we also use the
{\bibless}~\citep{kiela2015exploiting} and {\hyperlex}~\citep{vulic2016hyperlex}
datasets. We refer to \citet{mine/roller2018hearst} for an in-depth discussion
of these datasets.

Table~\ref{tab:results} shows the results for all tasks on these datasets. It can be
seen that our proposed approach provides substantial gains on the detection and
directionality tasks and, overall, achieves state of the art results on seven of
these nine benchmarks. In addition, our method clearly outperforms other
embedding-based approaches on {\hyperlex}, although it can not fully match the
performance of the count-based methods. As \citet{mine/roller2018hearst} noted,
this might be an artifact of the evaluation metric, as count-based methods
benefit from their sparse-predictions in this setting.

It can also be seen that our method outperforms Poincar\'{e} \textsc{GloVe} for the
task of hypernymy prediction. While \citet{tifrea2018poincar} report $0.341$
Spearman's $\rho$ on \textsc{HyperLex} and $0.652$ accuracy on \textsc{WBless},
our method can achieve substantially better results for the same tasks ($\rho =
0.59$ on \textsc{HyperLex}, $\text{ACC} = 0.909$ on \textsc{WBLess}). This
illustrates the importance of the distributional constraints that are provided
by the Hearst patterns.

An additional benefit is the efficiency of our embedding. For all tasks, we have
used a 20-dimensional embedding for \textsc{HypeCones}, while the best results
for SVD-based methods have been achieved with 300 dimensions. This reduction in
parameters by over an order of magnitude clearly highlights the efficiency of
hyperbolic embeddings for representing hierarchical structures.

\paragraph{Reconstruction}
In the following, we compare embedding and pattern-based methods on the task of
reconstructing an entire subtree of \textsc{WordNet}, i.e., the animals, plants,
and vehicles taxonomies, as proposed by \citet{kozareva2010semi}. In addition to
predicting the existence of single hypernymy relations, this allows us to
evaluate the performance of these models for inferring full taxonomies and to
perform an ablation for the prediction of missing and transitive relations.
We follow previous work
\citep{kgs/bordes2013translating,mine/nickel2017poincare} and report for each
observed relation $(u, v)$ in \textsc{WordNet}, its score ranked against the
score of the ground-truth negative edges. In Table~\ref{tab:hovy}, \emph{All} refers
to the ranking of all edges in the subtree, \emph{Missing} to edges that are not
included in the Hearst graph $G$, \emph{Transitive} to missing transitive edges
in $G$ (i.e. for all edges ${\{((x, z) : (x, y), (y, z) \in E \land (x, z)
  \notin E \}}$).

It can be seen that our method clearly outperforms the SVD and count-based
models with a relative improvement of typically over $40\%$ over the best
non-hyperbolic model. Furthermore, our ablation shows that \textsc{HypeCones}
improves the consistency of the embedding due to its transitivity property. For
instance, in our Hearst Graph the relation (\emph{male horse}, \isa,
\emph{equine}) is missing. However, since we correctly model that (\emph{male
  horse}, \isa, \emph{horse}) and (\emph{horse}, \isa, \emph{equine}), by
transitivity, we also infer (\emph{male horse}, \isa, \emph{equine}), which SVD
fails to do.


\section{Conclusion}
\label{sec:conclusion}
In this work, we have proposed a new approach for inferring concept hierarchies
from large text corpora. For this purpose, we combine Hearst patterns with
hyperbolic embeddings what allows us to set appropriate constraints on the
distributional contexts and to improve the consistency in the embedding space.
By computing a joint embedding of all terms that best explains the extracted
Hearst patterns, we can then exploit these properties for improved hypernymy
prediction. The natural hierarchical structure of hyperbolic space allows us
also to learn very efficient embeddings that reduce the required dimensionality
substantially over SVD-based methods. To improve optimization, we have
furthermore proposed a new method to compute entailment cones in the Lorentz
model of hyperbolic space. Experimentally, we show that our embeddings achieve
state-of-the-art performance on a variety of commonly-used hypernymy benchmarks.


\bibliography{paper_bibtex.bib}
\bibliographystyle{acl_natbib}

\end{document}